# On the Use of Different Feature Extraction Methods for Linear and Non Linear kernels

Imen TRABELSI[*], Dorra BEN AYED [*,**]

[*]*Institute of Computer Science of Tunis (ISI), Tunis, Tunisia*
trabelsi.imen1@gmail.com

[**]*National School of Engineer of Tunis (ENIT), Tunis, Tunisia*
Dorra.mezghani@isi.rnu.tn
dorrainsat@yahoo.fr

**Abstract:** The speech feature extraction has been a key focus in robust speech recognition research; it significantly affects the recognition performance.
In this paper, we first study a set of different features extraction methods such as linear predictive coding (LPC), mel frequency cepstral coefficient (MFCC) and perceptual linear prediction (PLP) with several features normalization techniques like rasta filtering and cepstral mean subtraction (CMS).
Based on this, a comparative evaluation of these features is performed on the task of text independent speaker identification using a combination between gaussian mixture models (GMM) and linear and non-linear kernels based on support vector machine (SVM).

## INTRODUCTION

In this paper, we highlight some of key related researches, techniques and approaches that have arisen to extract the suitable feature parameters. Currently, there are two major approaches to feature extraction: modeling human voice production and modeling perception system. In the first model, the voice evolved primarily to produce speech for conversation, in the second model, hearing evolved to recognize these sounds. So we try to classify the features extraction under these two models.

In order to enhance performance and robustness in automatic speech recognition, pre processing and filtering in speech feature extraction are commonly used.

In this paper, we motivate the use of extraction feature techniques for text independent speaker identification system using the GMM supervector in a support vector machine (SVM) classifier.

## 1. Different speech feature

Features extraction in ASR is the computation of a sequence of feature vectors which provides a compact representation of the given speech signal.

Producing and perceiving speech are basin human activities, a speaker can be presented as an encoder in a speech production process and the listener can be presented as a decoder in a speech perception process.

Figure 1 shows the complete process of producing and perceiving speech from the formulation of a message of a talker, to the creation of the speech signal, and finally to the understanding of the message by a listener.

Between human auditory and speech production systems, some researches believe that the auditory system came first, other researches uses the speech production model as the primary focus [URS 02].

It is the acoustic speech signal which mediates between the two systems. Thus, it is only natural to expect that the properties of the acoustic signal can tell us about both the human speech production system and the human auditory system.

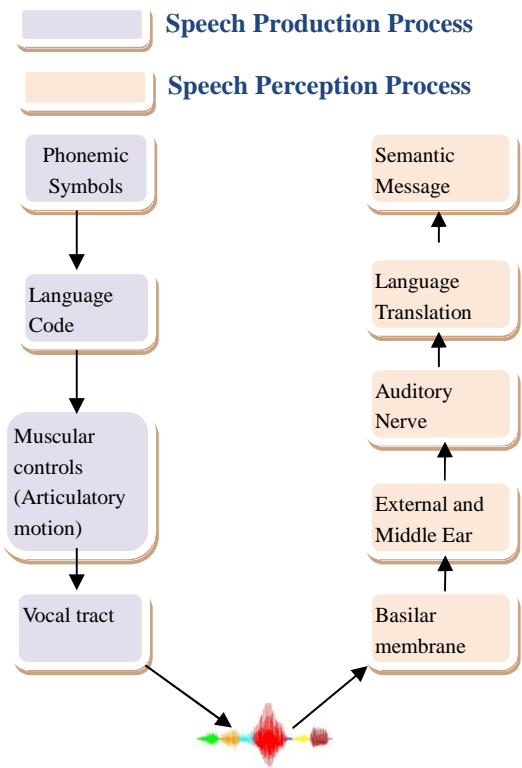

**Figure 1.** *Speech Production/Perception Model*

This section reviews the production/perception process and the discriminates features extracted from their characteristics.

**1.1. Features based on speech production**

We elucidate in this section the production mechanisms that give rise to different kinds of features.

Speech production is produced by the combined motion of articulatory gestures.

The mechanism of speech is composed of four processes: language processing, in which the content of an utterance is represented somehow in the brain; generation of motor commands to the vocal organs; articulatory movement for the production of speech by the vocal organs based on these motor commands; and the emission of air sent from the lungs in the form of speech [HON 03].

These structures are able to generate and shape a wide variety of waveforms. These waveforms can be broadly categorized into voiced and unvoiced speech.

These features describe properties of speech production rather than the properties of the acoustic signal resulting from it.

Based on the knowledge of the speech production mechanisms, we are able to extract a set of features which can best represent a particular phoneme. The phonemes are classified in terms of manner of articulation, (how is the vocal tract constricted), and place of articulation (where is the vocal tract constricted?) as mentioned in the table below.

**Table 1.** *Places and manners of articulation*

| Feature | Value |
|---|---|
| Manner of articulation | Voiced |
| | Invoiced |
| | Consonantal |
| | Nasal |
| | Sonorant |
| Place of articulation | Labial |
| | Dental |
| | Alveolar |
| | Palatal |
| | Velar |

Some features are motivated from a speech point of view like articulatory features and linear predictive analysis.

**1.1.1.** *Articulatory features*

Articulatory features (AFs) have attracted interest from the speech recognition community for more than a decade for many reasons [KIN 07] [KIR 00].

These features describe the configuration of the human vocal tract and the properties of speech production.

The basic idea of this approach is to bears an affinity to the articulatory events underlying the speech signal.

This representation is composed of classes describing the most essential articulatory properties of speech sounds such as place, manner, voicing, lip-rounding, the opening between the lips, and the position of the tongue.

**1.1.2.** *Linear Predictive Coding - LPC*

Linear predictive analysis of speech were introduced in late 1960s and become the predominant technique for estimating the basic parameters of speech [MAK 75].

Based on a highly simplified model for speech production, LPC provides both an accurate estimate of the speech parameters such as pitch, formants and spectra. It tries to imitate the human speech production mechanism. In addition, all the vocal tract parameters are represented in a set of LPC coefficients. The number of coefficients is typically 10 to 20 [KIN 07].

It is widely used because it is fast, simple and its ability to extract and store time varying formant information.

**1.2. Features based on perception system**

The auditory system is the sensory system for the sense of hearing. Research in speech perception seeks to understand how human listeners recognize speech sounds and use this information to understand spoken language.

The acoustic wave is transmitted from the outer ear to the inner ear which performs a transduction from acoustic energy to mechanical vibrations which ultimately are transferred to the basilar membrane inside the cochlea (the main component of the inner ear). The mechanical vibrations are then transuded into activity signals on the auditory nerves corresponding to a feature extraction process [KIM 99].

The cochlea performs the filterbank based frequency analysis on the speech signal to extract the pertinent features. Thus, most techniques are pivoting around the filterbank methodology in extracting the features. The difference in the design of the filterbank offers the extraction of different features from the signal.

In fact the question of the imitation of the human auditory system characteristics for ASR has been subject of discussion and some researches believe that the analysis based on the effective peripheral auditory processing is the most robust front end in noise [HER 98].

From the point of view of speech perception, we can describe some of these features.

**1.2.1.** *Mel Frequency Cepstral Coefficients - MFCC*

The most commonly used acoustic features are Mel-scale frequency cepstral coefficient based on frequency domain using the Mel scale which is based on the human ear scale.

MFCC takes human perception sensitivity with respect to frequencies into consideration.

MFCC is based on psychoacoustic research on the pitch and the perception of different frequency bands by the human ear. These parameters are similar to ones that are used by humans for hearing speech.

**1.2.2.** *Perceptual Linear Prediction -PLP*

Hermansky [HER 90] introduced a new technique, perceptual linear predictive (PLP) analysis.

This technique is based on the short-term spectrum of speech. It combined several engineering approximations to selected characteristics of human hearing and approximates auditory spectra by an autoregressive all-pole model.

PLP uses engineering approximations for three basic concepts from the psycho-acoustic of hearing: spectrum critical band spectral resolution, the equal-loudness curve and intensity power low.

Like MFCC, PLP employ an auditory based warping of the frequency axis derived from the frequency sensitivity of human hearing.

### 1.3. Other speech features

**1.3.1.** *Dynamic features*

The set of features described so far capture the average frequency distribution during a frame. Important information in the speech signal is however contained in the temporal evolution of the signal, in its dynamics.

One way to capture this information is to use the dynamic properties of speech, the first and/or second order differences of static coefficients which are called the delta (speed) and delta-delta (acceleration) coefficients. The time derivative is approximated by differencing between frames after and before the current, for instance:

$$\Delta_{yi} = y_{i+d} - y_{i-d} \qquad (1)$$

Where $y^i$ is the feature vector at frame i, and d typically is set to 1 or 2.

It has become common to combine dynamic features with the basic static features. It usually results in better performance.

**1.3.2.** *Prosodic features*

Prosody is defined as any property of speech that is not limited to a specific phoneme.

Prosody is a term that refers to the suprasegmental aspects of speech, including variations in pitch (fundamental frequency), energy, loudness, duration, pause, intonation, rate, stress and rhythm.

Prosody may reflect various features of the speaker, his emotional state or speaking style.

Very few people have done experiments which directly incorporate prosody as complementary information with ASR.

Kompe [KOM 97] is one of the few people to experiment with prosody. He reports improvements to recognition rates when prosodic information is used for recognition purposes.

## 2. Pre-processing

The speech preprocessing part is the fundamental signal processing applied before extracting features from speech signal, conditions the raw speech signal and prepares it for subsequent manipulations. Commonly used preprocessing techniques are illustrated as follows.

### 2.1. Digitalization

It is the first step in the speech processing speech acquisition, requires a microphone coupled with an analog-to-digital converter to receive the voice speech signal, sample it, and convert it into digital speech.

The analog speech signal is digitized with sampling rate of 8 KHZ in digital telephony and 10 KHZ, 12 KHZ or 16 KHZ in non telecommunication application.

### 2.2. End Point Detection

This step is based on signal energy evaluation. A voice signal can be divided into three parts: speech segment, silence segment and background noise. In order to segregate between them we call algorithms for speech end point detection. After which the unnecessary parts have been removed.

### 2.3. Pre-emphasis

The digitized speech signal Y[n] is sent to a Finite Impulse Response (FIR) Filter:

$$Y[n] = x[n] - \alpha x[n-1] \quad (2)$$

$$0.9 \leq \alpha \leq 1.0 \quad (3)$$

Where x[n] is the input speech signal and Y[n] is the output pre-emphasized signal and α is an adjustable parameter.

The goal of pre-emphasis is to compensate the high-frequency part that was suppressed during the sound production mechanism of humans. Moreover, it can also amplify the importance of high-frequency formants.

### 2.4. Frame blocking

The continuous Pre-emphasis signal Y is divided into overlapping frames of N samples.

Frame duration typically rages between 10 and 30 ms short time intervals to guarantee the quasi-stationary of the signal with optional overlap of [1/3 1/2] of the frame size.

### 2.5. Windowing

After framing, windowing techniques are applied in order to reduce the effect of discontinuity in every frame and at the edges of the frame.

Each frame has to be multiplied with a windowing technique, there are different types of windowing functions, like rectangular, hamming, barlett, Blackman, Kaiser, bohman, chebyshev, hanning and gaussian windows. The most popular is the hamming window w (n), is defined by:

$$w(n,a) = (1-a) - a\cos(2\Pi n / N-1) \quad (4)$$

$$0 \leq n \leq N-1 \quad (5)$$

Different values of a corresponds to different curves for the Hamming windows

Then the signal in a frame S[n] after Hamming windowing is:

$$S[n] = Y[n] * w[n] \quad (6)$$

## 3. Post Processing

This section reviews the various methods which have been proposed for feature normalization.

### 3.1. Cepstral Mean Subtraction -CMS

The algorithm computes a long-term mean cepstral value of the feature vectors and subtracts the mean value from the cepstral vectors of that utterance and then produces a normalized cepstrum vector. CMS avoids the low frequency noise to be further, amplified but the average vocal tract configuration information pertaining to the speaker may be also lost [FUR 81].

### 3.2. Cepstral variance normalization -CVN

Cepstral variance normalization is also known as the mean and variance normalization (MVN) [JAI 01] because CVN is often used in conjunction with CMS. The mean and variance of cepstral coefficients are assumed to be invariant in the CVN analysis. Therefore the exclusion of these properties would result in the removal of irrelevant information such as the effects of mismatched environments.

### 3.3. RASTA-filtering

Rasta-filtering was proposed for robust speech recognition by Hermansky and Morgan [HER 94].

At the beginning, it was introduced to support Perceptual Linear Prediction (PLP) preprocessing. It uses bandpass filtering in the log spectral domain. It has also been applied to other cepstral feature based preprocessing with both log spectral and the cepstral domain filtering.

Rasta filtering then removes slow channel variations and makes PLP more robust to linear spectral distortions [HER 91]. This technique has proven to be a successful technique for channel normalization in automatic speech recognition.

### 3.4. Feature warping

Also known as cumulative distribution mapping [CHOI 06]. It consists of mapping the observed cepstral feature distribution to a predefined target distribution over a sliding window with zero mean and unit variance [PEL 03].

This technique is a real-time equivalent of histogram equalization in image processing that maps a source feature distribution to a target distribution. This feature processing technique has successfully been applied to speaker recognition because it is robust to channel mismatch, additive noise and to some extent, nonlinear effects attributed to handset distortion [PEL 03] [BAR 03].

### 3.5. Short-time Gaussianization

This is achieved by an iterative scheme in each iteration; a global linear transformation is applied to the features in order to make them more dependant or decorelated in the new feature space before mapping them to an ideal distribution, such as the standard normal distribution [CHEN 00]. This linear transformation can be estimated by Expectation Maximization (EM) algorithm [DEM 77].

# 4. Experimental evaluation of different features with application in speaker identification

## 4.1. System Conditions

Our baseline system is a text-independent speaker identification task based on hybrid GMM/SVM system [BOU 09a], [BOU 09b].

The purpose of this study is to evaluate the performance of different acoustic features, when training data and testing data are in a CLEAN environment in order to show the differences between them.

Experiments have been conducted under the experimental conditions described (in table 2).

**Table 2.** *System Baseline*

| Corpus | Timit |
|---|---|
| Dialect | DR1 |
| Speaker | 14 Female |
| Number utterance per speaker for train | 8 sentences |
| Number utterance per speaker for test | 2 sentences |

In our experiments, we evaluated several different feature measurements, including Mel-scale Frequency Cepstral (MFCCs), Perceptual Linear Prediction (PLP), Linear Predictive Coding (LPC) both with and without their first and second derivatives and combined with normalization techniques.

The performance is measured as the identification rate ( IR).

$$IR (\%) = \frac{\text{Number correct assignments}}{\text{Number total assignments}} \times 100 \quad (7)$$

Specifications of the input audio stream at the acoustic pre-processor are summarized as follows (in table 3).

**Table 3.** *Pre-Processing Stages*

| Stage | Value |
|---|---|
| Sampling rate | 16 KHZ |
| End Point Detection | Energy based VAD |
| Pre Emphasis | $1 - 0.95 z^{-1}$ |
| Frame Duration | 16 ms |
| Frame Shift | 8ms |
| Windowing | Hamming |

## 4.2. GMM-UBM baseline System

The baseline system was a GMM UBM [REY 95]. Speaker modeling involves 2 step processes: a general universal background model UBM is trained using acoustic data of different speakers, in order to model the acoustics of speech.

The UBM comprised of 128 mixtures is trained using the EM algorithm with a vector quantization pre-estimate (KMEANS). And a target speaker model is adapted by Bayesian adaptation MAP [REY 00] from the UBM by adjusting the UBM means.

All Gaussian means vectors are pooled together to get one GMM Supervector [CAM 06]. The GMM supervector can be thought of as a mapping between an utterance and a high-dimensional vector. The Process is shown in Figure2.

We produced GMM supervector on a per utterance using MAP adaptation.

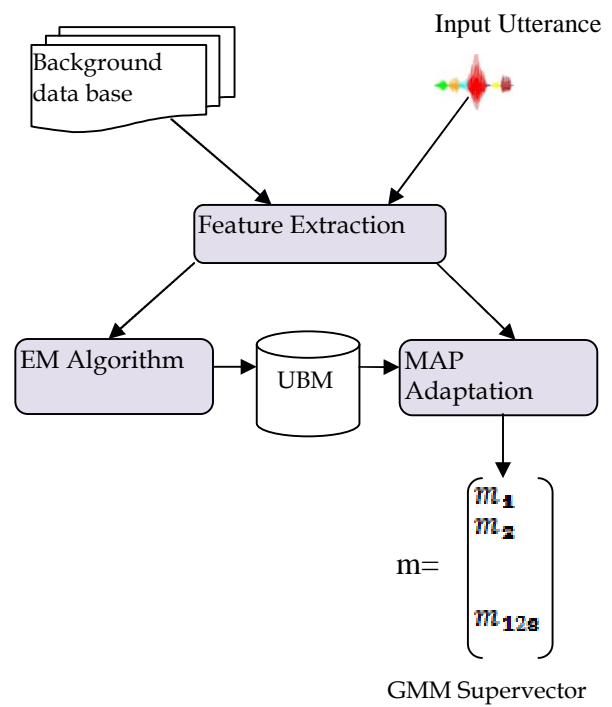

**Figure 2.** *Process of generating GMM Supervector*

## 4.3. Support vector machine in the GMM space

GMM-supervector and SVM combines both generative and discriminative methods and leads to the generative SVM kernels based on the probability distribution estimation.

In our case, SVM is applied in the GMMs means supervector space [SCH 96] as shown in figure 3.

SVMs perform a nonlinear mapping from an input space to an SVM expansion space.

The main design component in an SVM is the kernel, which is an inner product in the SVM feature space.

In our experiments, we have used two different kernel functions, the first one corresponds to the linear GMM-SVM kernel. The last ones is the non linear GMM-SVM kernels based on the radial basis function

(RBF).

The Kernel in equation (8), (9) and scoring are implemented using the library LIBSVM [CHA 01].

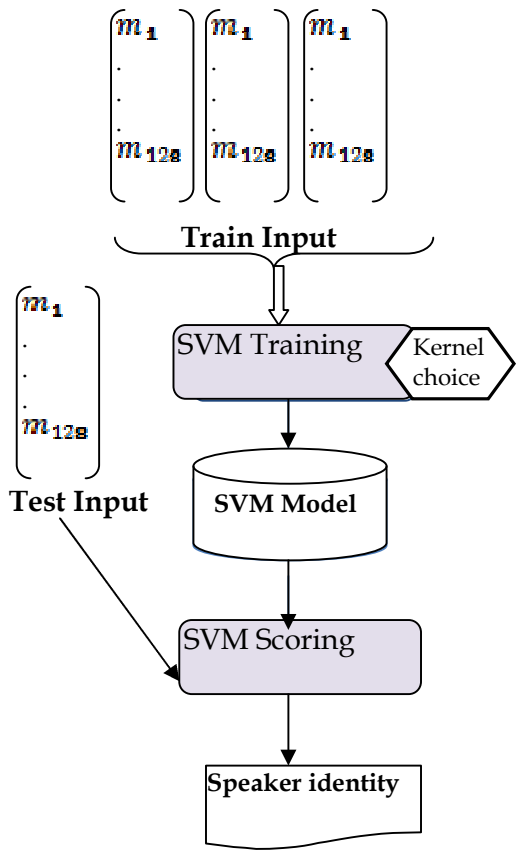

**Figure 3.** *SVM Design in the GMM space*

$$k(x, v_i) = x \cdot v_i \quad (8)$$

where x is the input data and vi are the support vectors.

$$K(x, v_i) = \exp[-\frac{1}{2\sigma}(x - v_i)^2] \quad (9)$$

where σ is the standard deviation of the radial basis function.

The best RBF parameters are chosen with a cross validation. We first divide the training set into 10 folds of equal size. Sequentially one fold is tested using the classifier trained on the remaining 9 folds.

Various parameters values are tried and the one with the best cross-validation accuracy is picked.

**4.4. Experiments results**

Our motivation was to analyze how much the performance rates for speaker identification task are depending on the choice of feature extraction techniques and on the kernel functions training SVMs.

*4.4.1. MFCCs variants Results*

Table 4 shows the performance of our GMM SVM system based on different combination of MFCC.

We observe that the IR is identical for MFCC, MFCC + delta, and MFCC + Energy. This IR is equal to 100% but when MFCC are combined with delta and delta delta, our system achieve an IR between 92, 85% and 96, 42% where the best performance in this case was reported by linear kernel.

When MFCC is used together with its delta, delta-delta and energy, IR obtained is around 96, 42% for both linear and RBF kernels.

Whereas IR degrades most significantly when MFCC are enhancing with CMS clearly with RBF kernel.

**Table 4.** *Results identification rate IR using various combinations of MFCC*

| Feature Type | Number | IR (%) Linear | RBF |
|---|---|---|---|
| MFCC | 12 | 100 | 100 |
| MFCC + Energy | 13 | 100 | 100 |
| MFCC +Δ | 24 | 100 | 100 |
| MFCC+ Δ+ ΔΔ | 36 | 96,42 | 92,85 |
| MFCC+Δ+ΔΔ+ Energy | 39 | 96,42 | 96,42 |
| MFCC+ CMS | 12 | 53,57 | 46,42 |

*4.4.2. PLP variants Results*

Table 5 presents the results obtained for different combination of PLP. These results show that IR keep the same value for PLP and PLP+ first delta+ second delta. It is equal to 100%. But decline significantly when PLP are combined with Rasta filter importantly with RBF kernel.

**Table 5.** *Identification rate using various combinations of PLP*

| Feature Type | Number | IR (%) Linear | RBF |
|---|---|---|---|
| PLP | 13 | 100 | 100 |
| PLP +Δ | 26 | 85,71 | 82,14 |
| PLP+ Δ+ ΔΔ | 39 | 100 | 100 |
| MFCC+Δ+ΔΔ+ RASTA filter | 39 | 96,42 | 53,57 |

*4.4.3. LPC variants Results*

In Table 6, we compare the performance of our system using LPC variants.

When LPC is associated with dynamic features, the IR increase from 96, 42% to 100% for linear and RBF kernels.

**Table 6.** *Identification rate using various combinations of LPC*

| Feature Type | Number | IR (%) Linear | RBF |
|---|---|---|---|
| LPC | 13 | 96,42 | 96,42 |
| LPC +Δ | 26 | 100 | 100 |
| LPC+ Δ+ ΔΔ | 39 | 100 | 100 |

## 5. Conclusion

A major problem in speech recognition system is the decision of the suitable feature set which can faithfully describe in an abstract way the original speech signal.

The objective of this paper was to give account of the current knowledge on the area of features extraction, speech pre processing and normalization methods for speaker identification tasks.

We have proposed an hybrid GMM-SVM system. We have presented the performance of this combination with different features and two different Kernels.

Then a comparative study was made to investigate the best choice of kernel function and the best input features.

First, we conclude that MFCC and LPC outperform PLP.

We also conclude that including the delta and acceleration coefficients has a negative affect on ASR performance excluding with LPC.

We therefore conclude that Rasta Filter and CMS did not improve accuracy.

This happens because Data in TIMIT were recorded with high-quality desktop microphones in a clean environment and does not include session variability between train and test. In this case, temporal coefficients and normalization methods remove useful information.

In addition, our experiments reveals that linear and RBF kernels give equal performance with a small favor for linear SVM.

Thus, as a future work, we will try to study the performance of SVM for speaker identification task by using all dialects of the TIMIT corpus and eventually extend our study to other different environments with acoustic mismatch. And we will attempt to study the performance of other SVM kernels.